\documentclass[runningheads]{llncs}
 
\usepackage{eccv}

\usepackage{graphicx}
\usepackage{booktabs}
\usepackage[accsupp]{axessibility}
\usepackage[table]{xcolor}

\usepackage{hyperref}
\usepackage{orcidlink}

\definecolor{lightblue}{RGB}{120,170,255}
\definecolor{lightgreen}{RGB}{140,210,150}
\definecolor{lightgrey}{RGB}{150,150,150}
\definecolor{lightorange}{RGB}{255,190,120}
\definecolor{highlight}{RGB}{230,230,230}

\begin{document}

% ---------------------------------------------------------------
\title{BeyondSight: Object Permanence for \\ End-to-End Autonomous Driving} 

% TODO FINAL: Replace with your author list. 
% Include the authors' OCRID for the camera-ready version, if at all possible.
\author{Sandro Papais\inst{1} \and
Letian Wang\inst{1} \and
Mudit Jain\inst{2} \and
Behnaz Rezaei\inst{2} \and
Steven L. Waslander\inst{1}
}

% TODO FINAL: Replace with an abbreviated list of authors.
\authorrunning{S.~Papais et al.}
% First names are abbreviated in the running head.
% If there are more than two authors, 'et al.' is used.

% TODO FINAL: Replace with your institution list.
\institute{University of Toronto \\
\email{\{sandro.papais, letian.wang, steven.waslander\}@robotics.utias.utoronto.ca} \and
Automated Driving, Qualcomm Technologies, Inc.\\
\email{\{mudijain, brezaei\}@qti.qualcomm.com} 
}

\maketitle

\vspace{-0.8em}
\begin{center}
\small Project page: \url{https://beyondsight-eccv.github.io}
\end{center}
\vspace{-0.8em}

% ---------------------------------------------------------------
\begin{abstract}
  % The abstract should concisely summarize the contents of the paper. 
  % While there is no fixed length restriction for the abstract, it is recommended to limit your abstract to approximately 150 words.
  % Please include keywords as in the example below. 
  % This is required for papers in LNCS proceedings.
Autonomous driving operates in partially observable environments where actors may become fully occluded by other vehicles or infrastructure. 
Most end-to-end driving systems implicitly couple actor existence to instantaneous observations, causing actor hypotheses to degrade or disappear during prolonged occlusion and removing potentially critical agents from downstream prediction and planning.
We introduce \textbf{BeyondSight}, a permanence-aware end-to-end driving framework that decouples actor existence from observability by maintaining persistent actor hypotheses over time. 
BeyondSight propagates actor queries temporally and updates them with observation-conditioned evidence, enabling joint perception, prediction, and planning to reason about actors even when they are temporarily unobservable.
To enable principled training and evaluation of persistence-aware models, we further introduce \textbf{nuScenes-Permanence}, an extension of nuScenes that provides supervision and observability-conditioned evaluation for unobservable actors.
Experiments show that BeyondSight substantially improves reasoning under occlusion, increasing detection performance for unobservable actors from 0 to 0.249 mAP while reducing planning error from 0.61 to 0.54 L2$_{\text{avg}}$.
These results highlight \emph{object permanence} as an important modeling principle for robust end-to-end autonomous driving. 

\end{abstract}

% ---------------------------------------------------------------
\section{Introduction}
\label{sec:intro}

% Motivation
Autonomous driving operates in partially observable environments where actors frequently become fully unobservable due to occlusion or sensor limits. 
Our analysis of nuScenes shows that approximately 30\% of actors are fully unobservable at each timestep, including many within close proximity to the ego vehicle. 
Therefore, safe driving requires maintaining persistent hypotheses about actors even when they are temporarily unobservable.
This capability corresponds to \emph{object permanence}: the ability to reason about objects that continue to exist despite missing observations.

\begin{figure}[!htb]
    
    \centering
    \begin{subfigure}[t]{0.9\linewidth}
        \centering
        \caption{Existing observable driving stacks.}
        \includegraphics[width=\linewidth]{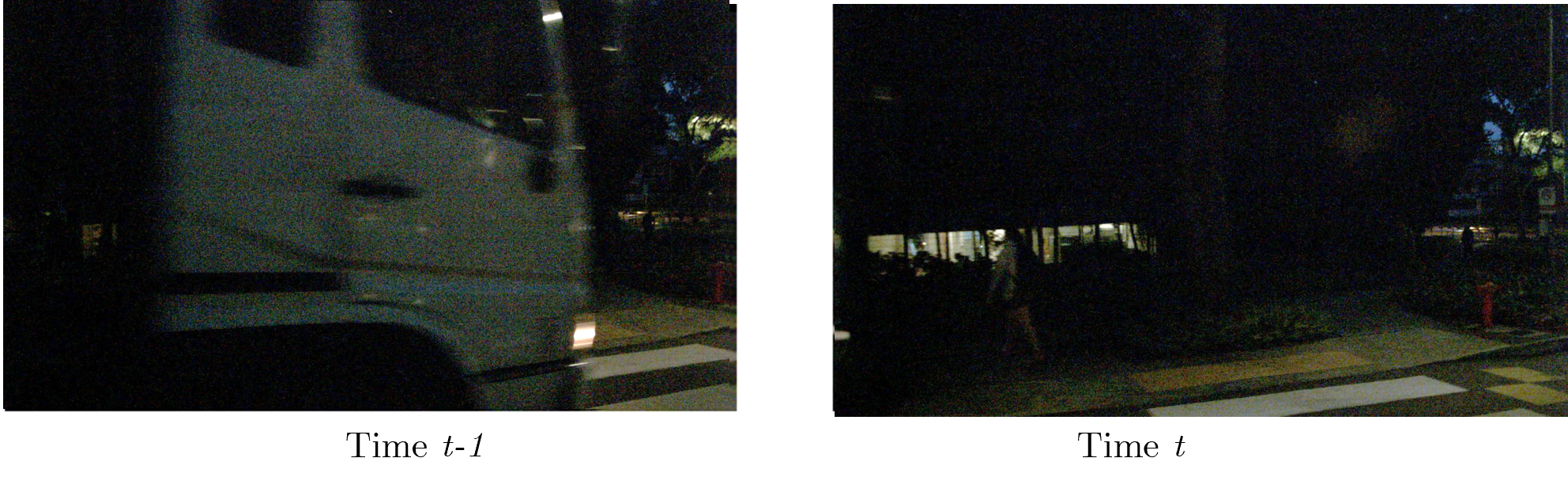}
    \end{subfigure}
    \begin{subfigure}[t]{0.9\linewidth}
        \centering
        \caption{BeyondSight.}
        \includegraphics[width=\linewidth]{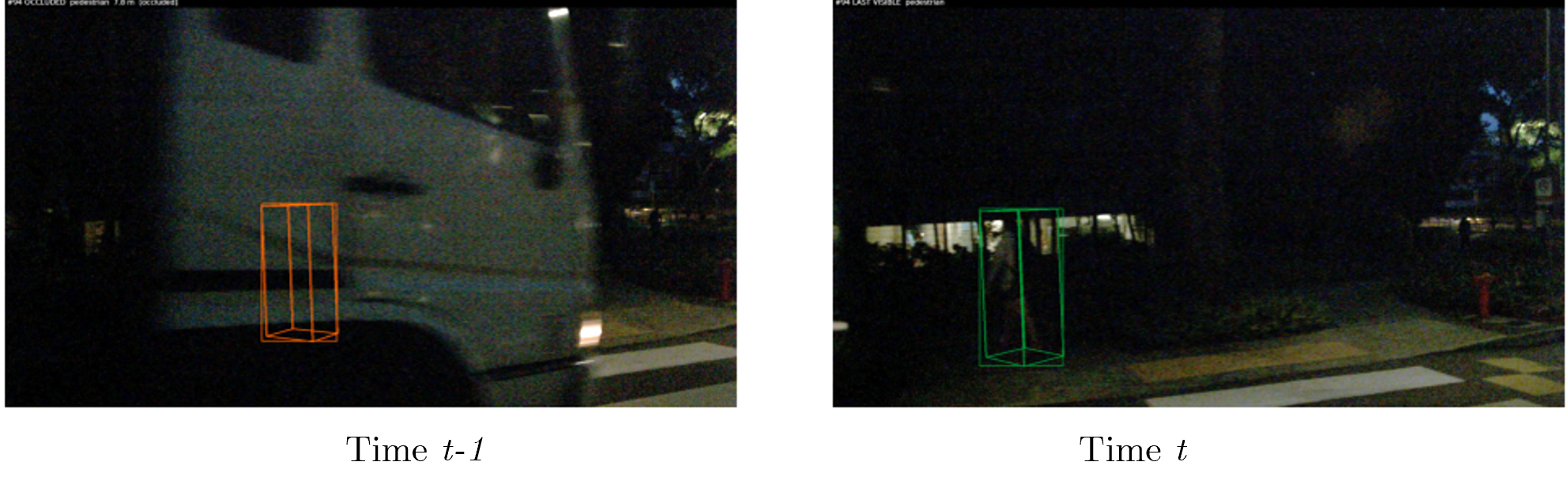}
    \end{subfigure}
    \caption{
    \textbf{Object permanence in prolonged occlusion.}
    A pedestrian near a crosswalk becomes occluded.
    Top: End-to-end driving models (ex. SparseDrive) drop the actor hypothesis.
    Bottom: BeyondSight maintains a persistent representation during occlusion.
    }
    \label{fig:intro_teaser}
    
\end{figure}

% Existing work limitations
Recent end-to-end autonomous driving (E2E-AD) systems jointly model perception, prediction, and planning within a unified architecture.
However, most existing approaches implicitly couple actor existence to instantaneous observations.
When an actor becomes fully unobservable, its representation often degrades or disappears entirely (Fig.~\ref{fig:intro_teaser}), removing it from the scene representation used by downstream prediction and planning modules.
As a result, the planner must reason only over currently visible agents, leading to reactive or brittle behavior in occlusion-heavy scenarios such as intersections and crosswalks.
While temporal aggregation improves short-term stability, it does not explicitly model actor persistence during extended observability gaps.

% Method
Current perception benchmarks implicitly equate observability with existence: when an actor becomes fully occluded, it typically disappears from both supervision and evaluation.
This prevents models from learning persistent representations across observability gaps.
To address this limitation, we introduce \textbf{nuScenes-Permanence}, an extension of nuScenes that provides annotations and evaluation protocols for unobservable actors, enabling systematic training and evaluation of object permanence.

Building on this benchmark, we introduce \textbf{BeyondSight}, a permanence-aware end-to-end driving framework that explicitly decouples actor existence from instantaneous observability by maintaining persistent actor hypotheses over time.
BeyondSight extends sparse-query scene representations with temporal propagation and observation-conditioned updates, allowing actor hypotheses to persist through observability gaps.
Propagated and observation-conditioned hypotheses are fused into a unified representation that contains both observable and unobservable actors.
This persistent scene representation provides richer context for downstream motion prediction and planning, improving decision-making in occlusion-heavy scenarios.

% Results / Contributions
Our contributions are summarized as follows:
\begin{itemize}
\item We formalize \textbf{object permanence for end-to-end autonomous driving}, defining the requirement that actors previously observed remain represented even when temporarily unobservable in both model representations and dataset annotations.
\item We introduce \textbf{nuScenes-Permanence}, extending nuScenes with unobservable actor annotations and an observability-conditioned evaluation protocol that enables systematic training and evaluation of persistence-aware models.
\item We propose \textbf{BeyondSight}, a permanence-aware extension to sparse-query end-to-end driving architectures that maintains persistent actor hypotheses through temporal propagation and observability-aware fusion.
\end{itemize}

Extensive experiments on the nuScenes benchmark demonstrate the effectiveness of permanence-aware reasoning.
On nuScenes, BeyondSight improves planning error from 0.61 to 0.54 L2$_{avg}$ while maintaining competitive perception performance. On the proposed nuScenes-Permanence benchmark, it dramatically improves reasoning about occluded actors, increasing $\text{mAP}_\text{unobs}$ from 0 to 0.249.
Qualitative results further show that the model maintains stable hypotheses for actors that become fully occluded and produces safer planning behavior in complex scenes with prolonged occlusions.

% ---------------------------------------------------------------
\section{Related Work}
\label{sec:related}

\subsubsection{Vision-Based End-to-End Autonomous Driving.}
Vision-based end-to-end autonomous driving (E2E-AD) jointly optimizes perception, motion forecasting, and planning within a single differentiable stack.
Most modern E2E-AD systems operate in BEV, combining camera-to-BEV lifting with temporal aggregation to support long-horizon planning.
Lift-Splat-Shoot~\cite{philion2020lss}, ST-P3~\cite{hu2022st}, and BEVFormer~\cite{li2022bevformer} established the core recipe of BEV scene encoding and recurrent or attention-based temporal fusion.
Building on this, UniAD~\cite{hu2023planning} unified detection, tracking, prediction, and planning in a planning-centric architecture, while VAD~\cite{jiang2023vad} introduced vectorized scene representations that improve efficiency and downstream controllability.
Robust motion prediction remains a central requirement for deployable autonomy, especially under uncertainty, distribution shift, and interaction-heavy driving scenarios~\cite{wang2026trends}.

Recent work targets scalability and latency by replacing dense BEV supervision with sparse instance-centric representations.
SparseDrive~\cite{sun2025sparsedrive} uses sparse queries to represent agents and map elements for joint detection, prediction, and planning.
SSR~\cite{li2024navigation} further prunes supervision and computation via navigation-guided sparse tokens, while DriveAdapter~\cite{jia2023driveadapter} and DriveTransformer~\cite{jia2025drivetransformer} reduce coupling and improve streaming behavior through teacher--student decoupling and unified query processing.
Despite architectural progress, most E2E-AD stacks still rely on training signals and metrics defined over \emph{currently observable} actors by filtering annotations with zero LiDAR points, effectively removing fully occluded actors from supervision and evaluation.
As a result, actor hypotheses often degrade or disappear after missed detections or prolonged occlusion, leading to weak temporal consistency and limited recovery once the actor reappears.

\subsubsection{Temporal Modeling and Query Propagation.}
Temporal reasoning in vision-based driving is commonly implemented as (i) BEV feature recurrence or (ii) object-centric query propagation.
BEVFormer~\cite{li2022bevformer} aggregates historical BEV features with spatiotemporal attention, improving temporal stability for BEV reasoning.
In object-centric pipelines, StreamPETR~\cite{wang2023exploring} propagates instance queries across frames to enable efficient temporal association and multi-view 3D detection.
More recent E2E-AD systems extend these ideas to multi-task temporal coherence:
BridgeAD~\cite{zhang2025bridging} leverages historical prediction to inform present perception and planning, MomAD~\cite{song2025dont} introduces momentum objectives to stabilize planning, and ForeSight~\cite{papais2025foresight} shares memory between detection and forecasting to improve streaming consistency.
Related multi-object tracking methods also study hypothesis persistence through missed observations and association ambiguity. 
SWTrack~\cite{papais2024swtrack} maintains multiple 3D tracking hypotheses in a sliding window, while SCATr~\cite{cheong2026scatr}  mitigates query suppression for joint detection and tracking.

These approaches improve short-horizon stability, association, and interaction modeling, but query survival is still largely driven by recent observations and evaluation is typically defined only where ground truth remains observable.
As a result, hypotheses for fully unobservable actors often decay or disappear, yielding missing forecasts and incomplete context for planning.
BeyondSight instead treats persistence as a first-class requirement: temporally propagated hypotheses remain valid independent of immediate sensor evidence, and are explicitly supervised and evaluated across observability gaps.

\subsubsection{Occlusion and Object Permanence.}
Object permanence under occlusion has been studied in video understanding and tracking, where learning to preserve identity and state through occlusions requires explicit memory and supervision.
OPNet~\cite{shamsian2020learning} and PermaTrack~\cite{tokmakov2021learning} show that tracking through occlusion benefits from objectives that encourage a persistent latent state.
Subsequent work further demonstrates that heavy occlusion demands mechanisms and losses beyond short-term feature aggregation~\cite{tokmakov2022object,van2023tracking}.

In autonomous driving perception, several methods improve robustness to \emph{partial} occlusion by strengthening features or reasoning about difficult conditions (e.g., CorrBEV~\cite{xue2025corrbev} and ReasonNet~\cite{shao2023reasonnet}).
However, in end-to-end driving systems, actors that become fully unobservable are typically excluded from supervision and evaluation under standard protocols.
Temporal propagation may occasionally recover actors after brief observation gaps, but persistence across longer occlusions is neither explicitly supervised nor evaluated.
BeyondSight closes this gap by combining (i) permanence-aware supervision for fully unobservable actors, (ii) architectural state propagation across observability gaps, and (iii) an observability-conditioned evaluation protocol on nuScenes-Permanence.

% ---------------------------------------------------------------
\section{Problem Formulation}
\label{sec:problem_formulation}

\subsubsection{Partial Observability.}
We define the \emph{observability} of an actor using a binary indicator 
$o_t^i \in \{0,1\}$ for actor $i$ at time $t$.
Following the nuScenes protocol, observability is approximated by sensor support:
\begin{equation}
    o_t^i \approx \mathbb{I}\!\left(n_{\text{pts},t}^i > 0\right),
    \label{eq:observability}
\end{equation}
where $n_{\text{pts},t}^i$ is the number of associated LiDAR and radar returns.
A scene exhibits \emph{partial observability} when an actor that was previously observable becomes fully unobservable due to occlusion or sensor resolution.

\subsubsection{Object Permanence.}
We define \emph{object permanence} as the requirement that actors previously observed remain represented even when temporarily unobservable.
Formally, a representation satisfies permanence if
\begin{equation}
\forall i,t:\;
\Big(o_t^i = 0 \;\land\; \exists\, t' < t : o_{t'}^i = 1\Big)
\;\Rightarrow\;
\exists\, r_t^i \in \mathcal{R}_t,
\label{eq:permanence}
\end{equation}
where $\mathcal{R}_t$ denotes the set of actor representations maintained at time $t$ and $r_t^i$ corresponds to actor $i$.
Equation~\ref{eq:permanence} expresses permanence as a representation-level constraint that decouples actor existence from instantaneous observation.

In a driving model, $\mathcal{R}_t$ corresponds to the maintained hypothesis set, requiring actor hypotheses to persist across observability gaps.
In a dataset, $\mathcal{R}_t$ corresponds to annotated actor states, requiring trajectories to remain defined during unobservable intervals.
BeyondSight enforces Eq.~\ref{eq:permanence} architecturally (Sec.~\ref{sec:method}), while the nuScenes-Permanence benchmark enforces it through supervision and evaluation (Sec.~\ref{sec:permanence_benchmark}).

% ---------------------------------------------------------------
\section{Method}
\label{sec:method}

\begin{figure}[!htb]
  \centering
  \includegraphics[width=0.8\linewidth]{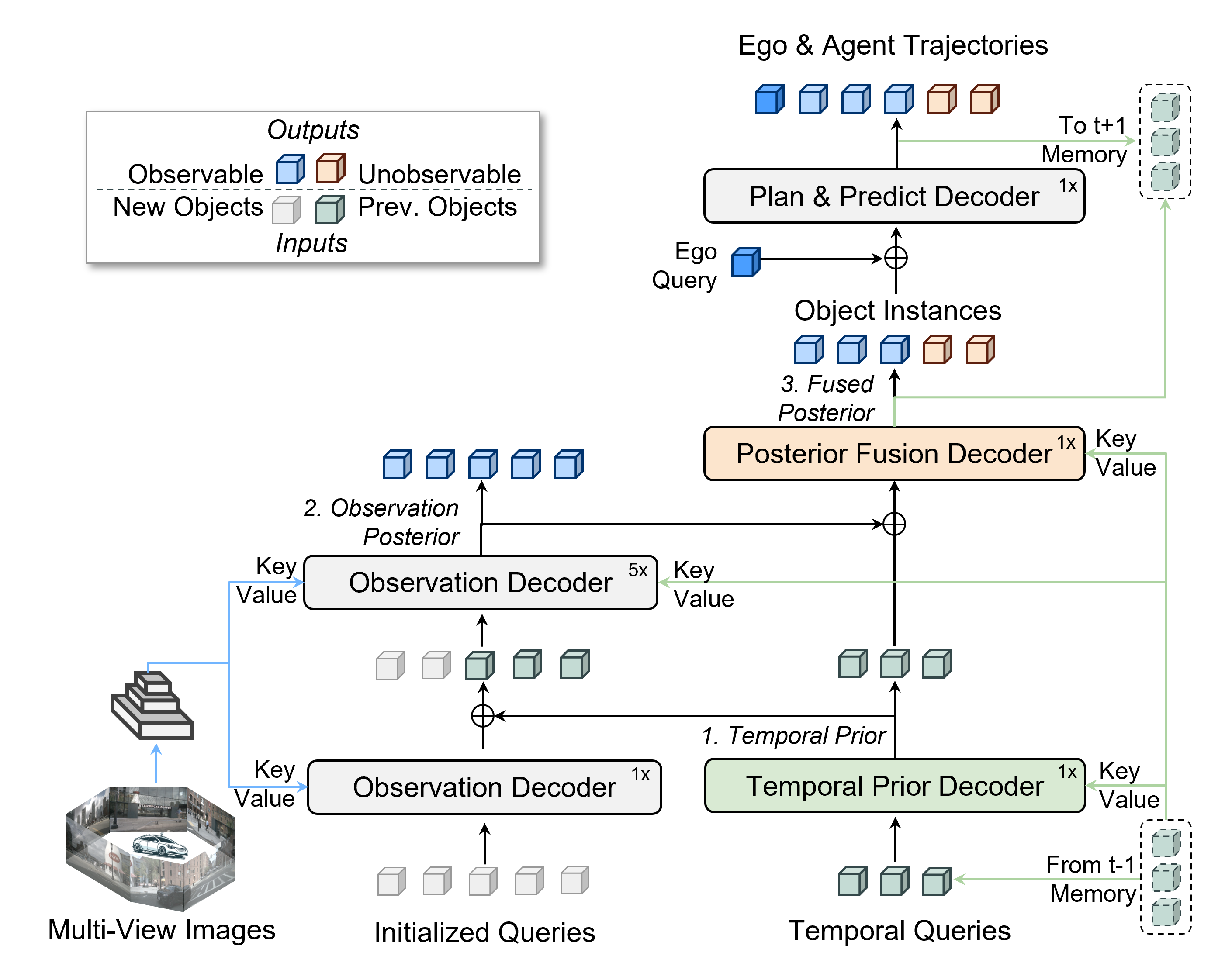}
\caption{BeyondSight overview. Actor belief update consists of three stages:
(1) a \textcolor{lightgreen}{\emph{Temporal Prior Decoder}} propagates actor hypotheses from the previous frame,
(2) an \textcolor{lightgrey}{\emph{Observation Decoder}} updates hypotheses using current image features,
and (3) a \textcolor{lightorange}{\emph{Posterior Fusion Decoder}} merges and refines propagated and observation-conditioned hypotheses.}
\label{fig:method}
\end{figure}

BeyondSight builds upon SparseDrive~\cite{sun2025sparsedrive}, a sparse query-based end-to-end driving architecture that jointly performs detection, prediction, and planning from multi-view camera inputs. The model maintains a set of instance queries representing dynamic actors and map elements, which are iteratively refined using deformable feature aggregation and temporal attention.

\subsection{BeyondSight Model}

To satisfy the actor persistence constraint defined in Sec.~\ref{sec:problem_formulation}, BeyondSight introduces a persistence-aware actor inference pipeline while retaining the SparseDrive backbone and objectives. Actor queries are updated through three stages resembling a Bayesian filtering approach:
\begin{align}
q_t^{\text{prior}} &= D_{\text{prior}}(q_{t-1}) \\
q_t^{\text{obs}} &= D_{\text{obs}}(F_t, q_t^{\text{prior}}) \\
q_t^{\text{post}} &= D_{\text{fusion}}(q_t^{\text{prior}}, q_t^{\text{obs}})
\end{align}
where $F_t$ denotes BEV features extracted from the perception backbone. The terms prior, observation, and posterior are used in a filtering-inspired sense: they denote learned latent query representations for temporal propagation, image-conditioned update, and fused refinement, rather than calibrated probability distributions.

This formulation separates temporal hypothesis propagation from observation updates, enabling actor representations to persist during observability gaps. Unlike standard sparse-query propagation, where previous detections are refined through the observation decoder and remain tied to recent visual evidence, BeyondSight propagates motion-conditioned hypotheses in an image-feature-free temporal prior decoder, then fuses them with observation-conditioned queries. This keeps unobservable actors available to prediction and planning without requiring the visual branch to hallucinate non-visible evidence.

\subsubsection{Temporal Prior Decoder.}
The Temporal Prior Decoder propagates actor queries from the previous frame to produce a motion-conditioned hypothesis set. Historical detection queries and motion-predicted queries are fused through a lightweight MLP and refined using transformer-style self-attention operating solely on the queries. This stage updates actor hypotheses based on temporal consistency without accessing image features:
\begin{equation}
q_t^{\text{prior}} = D_{\text{prior}}(q_{t-1}).
\end{equation}

\subsubsection{Observation Decoder.}
The temporal prior is combined with newly initialized object queries and passed to the standard SparseDrive detection decoder, which attends to BEV features to produce observation-conditioned hypotheses:
\begin{equation}
q_t^{\text{obs}} = D_{\text{obs}}(F_t, q_t^{\text{prior}}).
\end{equation}
This stage is supervised only using observable ground-truth actors, allowing the propagated prior to maintain hypotheses for actors that remain temporarily unobservable.

\subsubsection{Posterior Fusion Decoder.}
The Posterior Fusion Decoder reconciles propagated and observation-conditioned hypotheses. For each actor, the proposal with higher classification confidence is retained and refined through a final query refinement stage:
\begin{equation}
q_t^{\text{post}} =
D_{\text{fusion}}(q_t^{\text{prior}}, q_t^{\text{obs}}).
\end{equation}
The resulting actor set $q_t^{\text{post}}$ forms the scene representation used by the downstream motion prediction and planning modules. Unlike SparseDrive, which supervises only observable actors, BeyondSight also supervises actors that are temporarily unobservable, allowing occluded actors to influence trajectory prediction and planning.

\subsection{Optimization}

We follow the two-stage training protocol of SparseDrive. In Stage 1, the sparse perception module is trained to learn the scene representation. In Stage 2, perception, motion prediction, and planning modules are trained jointly end-to-end.
The overall objective is
\begin{equation}
\mathcal{L} =
\mathcal{L}_{\text{det}}
+ \mathcal{L}_{\text{det}}^\prime
+ \mathcal{L}_{\text{map}}
+ \mathcal{L}_{\text{motion}}^\prime
+ \mathcal{L}_{\text{plan}}
+ \mathcal{L}_{\text{depth}} .
\end{equation}

The standard detection loss $\mathcal{L}_{\text{det}}$ supervises the Observation Decoder using observable actors, while the modified detection loss $\mathcal{L}_{\text{det}}^\prime$ supervises the temporal and fusion decoders using both observable and unobservable actors.

\subsubsection{Permanence-Aware Supervision.}
Let $G_{\text{obs}}$ and $G_{\text{unobs}}$ denote observable and unobservable ground-truth actors at time $t$, with
\begin{equation}
G_{\text{all}} = G_{\text{obs}} \cup G_{\text{unobs}} .
\end{equation}
Motion supervision is extended to unobservable actors by applying the motion loss over the full actor set:
\begin{equation}
\mathcal{L}_{\text{motion}}^\prime =
\mathcal{L}_{\text{motion}}(G_{\text{all}}).
\end{equation}

To explicitly decouple existence from observability, the posterior decoder predicts an observability state $\hat{o}_t^i$ for each actor query. This is supervised using binary cross-entropy:
\begin{equation}
\mathcal{L}_{\text{obs}} =
\sum_{i \in G_{\text{all}}}
\mathrm{BCE}(\hat{o}_t^i, o_t^i).
\end{equation}

The final detection objective becomes
\begin{equation}
\mathcal{L}_{\text{det}}^\prime =
\mathcal{L}_{\text{det}}(G_{\text{all}})
+
\lambda_{\text{obs}} \mathcal{L}_{\text{obs}}.
\end{equation}

% ---------------------------------------------------------------
\section{Permanence Benchmark}
\label{sec:permanence_benchmark}

Standard perception benchmarks such as nuScenes~\cite{caesar2020nuscenes} evaluate only actors that are observable at each timestamp, typically defined by the presence of LiDAR or radar returns. 
When an actor becomes fully occluded or moves outside sensor range, annotations and evaluation are usually discontinued. 
As a result, current benchmarks implicitly equate \emph{observability} with \emph{existence}, preventing models from learning or being rewarded for maintaining actor hypotheses during observability gaps.
To support training and evaluation of persistence-aware models, we introduce \textbf{nuScenes-Permanence}, an extension of the nuScenes dataset that provides supervision for actors that remain present in the scene but are unobservable.
The benchmark extension targets approximately continuous actor motion, while abrupt hidden maneuvers that cannot be reliably inferred from surrounding observations are outside its scope.

\subsection{Annotation Extension.}
We extend actor trajectories through intervals where sensor support is absent by reconstructing missing states offline using physically grounded motion priors. 
Observable segments remain unchanged, while unobservable intervals are completed using interpolation between observed states or short-horizon extrapolation when actors leave the sensor field of view. 
Annotated 3D boxes with zero LiDAR or radar points are retained rather than filtered, enabling supervision for heavily occluded actors that remain present in the scene.

Applying this procedure expands the nuScenes annotations from 932k to 1.33M boxes (+30\%), transforming the dataset into an occlusion-aware benchmark that supports supervision across observability gaps. 
The trajectory completion model operates strictly offline and is not used at inference time. 
BeyondSight must therefore learn to maintain persistent hypotheses without access to completed trajectories.

To validate the generated annotations, we perform a holdout reconstruction study in which observable trajectory intervals are masked, reconstructed with the same offline pipeline, and compared against the original ground truth. 
We use the resulting reconstruction errors to calibrate an occlusion-duration-dependent matching tolerance for unobservable evaluation. 
Additional validation details are provided in the supplementary material.

\subsection{Observability-Conditioned Evaluation.}
Standard nuScenes evaluation penalizes predictions of unobservable actors as false positives, discouraging persistence across occlusion intervals. 
To evaluate object permanence, we introduce an \emph{observability-conditioned} protocol that partitions ground truth into observable ($G_{\text{obs}}$) and unobservable ($G_{\text{unobs}}$) subsets.
Each metric specifies a target ground-truth set $G$ and an ignored set $I$. 
Predictions are first matched to $G$ using the standard nuScenes matching procedure; remaining predictions matched to $I$ are removed prior to metric accumulation. 
This symmetric ignore rule prevents cross-penalization between observable and unobservable actors.
In addition to the standard detection metric $\text{mAP}$, we report:
\begin{itemize}
\item $\text{mAP}_{\text{obs}}$: performance on continuously observable actors,
\item $\text{mAP}_{\text{unobs}}$: performance during unobservable intervals,
\item $\text{mAP}_{\text{all}}$: performance across the full actor set.
\end{itemize}

Because reference trajectories for unobservable actors are partially extrapolated, we use an adaptive matching tolerance that accounts for accumulated motion uncertainty during unobservability. 
This tolerance reduces to the standard nuScenes threshold when the unobservability duration approaches zero, ensuring compatibility with the official evaluation protocol. 
Details of the trajectory completion model and tolerance estimation are provided in the supplementary material.

% ---------------------------------------------------------------
\section{Experiments}
\label{sec:experiments}

\subsection{Experimental Setup}
\label{sec:exp_setup}

\subsubsection{Dataset.}
We evaluate on nuScenes~\cite{caesar2020nuscenes}, a large-scale autonomous driving dataset containing 1{,}000 driving scenes of approximately 20\,s each.
Following standard practice, we use the official train/validation split (700/150 scenes).
Each keyframe provides synchronized multi-view imagery from six cameras together with 3D bounding box annotations at 2\,Hz for ten object classes.
In addition to the standard dataset, we evaluate on \textbf{nuScenes-Permanence}, our extended benchmark introduced in Section~\ref{sec:permanence_benchmark}.
The extension augments nuScenes with supervision for actors that remain physically present but become temporarily unobservable due to occlusion, sensor range limits, or field-of-view constraints.
This extension increases the number of annotated actor states by approximately 30\%, enabling systematic evaluation of persistence-aware models.

\subsubsection{Metrics.}
We report the standard nuScenes metrics to ensure comparability with prior work.
We evaluate \textit{detection} using mAP and NDS together with the standard nuScenes error metrics mATE, mASE, mAOE, mAVE, and mAAE~\cite{caesar2020nuscenes}.
\textit{Tracking} performance is evaluated using AMOTA, AMOTP, recall, and ID switches.
For \textit{motion forecasting} we report minADE, minFDE, Miss Rate (MR), and End-to-end Prediction Accuracy (EPA) following the VIP3D protocol~\cite{gu2023vip3d} and UniAD-style evaluation~\cite{hu2023planning}.
\textit{Planning} quality is measured using open-loop trajectory error and collision rate following the VAD evaluation protocol~\cite{jiang2023vad}.

To evaluate persistence under partial observability, we additionally report the proposed \textit{observability-conditioned metrics} on nuScenes-Permanence: $\text{mAP}_{\text{obs}}$, $\text{mAP}_{\text{unobs}}$, $\text{mAP}_{\text{all}}$.
These metrics isolate performance on observable and unobservable actors while applying symmetric ignore rules to prevent penalization.
For unobservable intervals whose reference trajectories are extrapolated, we use the adaptive matching tolerance described in Section~\ref{sec:permanence_benchmark}.

\subsubsection{Implementation Details.}
BeyondSight is implemented on top of SparseDrive and retains its backbone architecture, sparse scene representation, motion prediction, and planning modules.
Unless otherwise specified, all architectural hyperparameters follow the corresponding SparseDrive configuration.
The Temporal Prior Decoder and Posterior Fusion Decoder are implemented as a lightweight transformer decoder using self-attention, two-layer MLPs, and a detection head operating on the query embeddings, introducing less than $5\%$ additional parameters.

Training follows a two-stage protocol.
First, the sparse perception module is trained to learn the scene representation.
Second, perception, motion prediction, and planning modules are trained jointly in an end-to-end manner.
We extend detection and motion supervision to unobservable timestamps and introduce an observability classification loss with weight $\lambda_{\text{obs}}=1.0$.
Models are trained using AdamW with weight decay $10^{-3}$, cosine learning-rate scheduling, and 500 warm-up iterations. 
Stage 1 uses learning rate $4\times10^{-4}$, while Stage 2 uses learning rate $3\times10^{-4}$. 
The image backbone learning-rate multiplier is set to 0.5 in Stage 1 and 0.1 in Stage 2. 
Full optimization details are provided in the supplementary material.

\subsection{Main Result}

We compare BeyondSight to state-of-the-art vision-based end-to-end driving baselines that jointly model perception, forecasting, and planning.
Our primary baseline is SparseDrive~\cite{sun2025sparsedrive}, since BeyondSight directly extends its sparse-query representation and training protocol.
We additionally compare to representative end-to-end driving systems including UniAD~\cite{hu2023planning}, VAD~\cite{jiang2023vad}, and recent efficient E2E-AD approaches when supported by public code and evaluation settings.
All methods are evaluated under the same nuScenes splits and metrics; where necessary, we re-train models under a unified pipeline to ensure fair comparison.

\subsubsection{End-to-end Performance.}
We first report standard nuScenes leaderboard metrics to ensure comparability with prior work (Tables~\ref{tab:res_plan} and \ref{tab:res_perc}).
BeyondSight consistently improves the strongest SparseDrive baseline across multiple tasks while maintaining compatibility with the official evaluation protocol.

\begin{table}[!htb]
\centering
\caption{Open-loop planning performance on the nuScenes validation set. 
Lower L2 error and collision rate are better.}
\label{tab:res_plan}
\resizebox{1\linewidth}{!}{
\begin{tabular}{l|cccc|cccc}
\toprule
Method
& L2$_{1s}$ $\downarrow$
& L2$_{2s}$ $\downarrow$
& L2$_{3s}$ $\downarrow$
& L2$_{avg}$ $\downarrow$
& CR$_{1s}$ (\%) $\downarrow$
& CR$_{2s}$ (\%) $\downarrow$
& CR$_{3s}$ (\%) $\downarrow$
& CR$_{avg}$ (\%) $\downarrow$ \\
\midrule
ST-P3           & 1.33 & 2.11 & 2.90 & 2.11 & 0.23 & 0.62 & 1.27 & 0.71 \\
UniAD           & 0.48 & 0.96 & 1.65 & 1.03 & 0.05 & 0.17 & 0.71 & 0.31 \\
VAD-Base        & 0.41 & 0.70 & 1.05 & 0.72 & 0.07 & 0.17 & 0.41 & 0.22 \\
% DriveTransformer-L & 0.19 & 0.34 & 0.66 & 0.40 & 0.03 & 0.10 & 0.21 & 0.11 \\
SparseDrive     & 0.29 & 0.58 & 0.96 & 0.61 & 0.01 & 0.05 & 0.18 & 0.08 \\
BridgeAD        & 0.28 & 0.55 & 0.92 & 0.58 & 0.00 & 0.04 & 0.20 & 0.08 \\
MomAD           & 0.31 & 0.57 & 0.91 & 0.60 & 0.01 & 0.05 & 0.22 & 0.09 \\
% DiffusionDrive  & 0.27 & 0.54 & 0.90 & 0.57 & 0.01 & 0.02 & 0.04 & 0.02 \\
\rowcolor{highlight} BeyondSight
& \textbf{0.26} & \textbf{0.51} & \textbf{0.85} & \textbf{0.54}
& \textbf{0.01} & \textbf{0.04} & \textbf{0.16} & \textbf{0.07} \\
\bottomrule
\end{tabular}
}
\end{table}

\begin{table}[!htb]
\centering
\caption{Perception, tracking, and prediction performance on the nuScenes validation set. 
Best results are in bold.}
\label{tab:res_perc}
\begin{tabular}{lc|cc|c|cc}
\toprule
Method & Backbone 
& mAP $\uparrow$ 
& NDS $\uparrow$
& AMOTA $\uparrow$
& minADE $\downarrow$
& EPA $\uparrow$ \\
\midrule
% DriveTransformer-Large & & -- & -- & -- & -- & -- \\
% ViP3D \\
% VAD \\
% UniAD \\
SparseDrive            & ResNet50 & 0.415 & 0.526 & 0.372 & 0.610 & 0.492 \\
BridgeAD               & ResNet50 & 0.423 & 0.534 & 0.398 & 0.620 & 0.500 \\
MomAD                  & ResNet50 & 0.423 & 0.531 & 0.391 & 0.610 & 0.499 \\
\rowcolor{highlight} BeyondSight & ResNet50
& \textbf{0.427}
& \textbf{0.536}
& \textbf{0.401}
& \textbf{0.610}
& \textbf{0.502} \\
\bottomrule
\end{tabular}
\end{table}

On perception, BeyondSight improves detection from $0.415$ to $0.427$ mAP and from $0.526$ to $0.536$ NDS.
Tracking performance also improves, increasing AMOTA from $0.372$ to $0.401$.
Motion forecasting performance remains competitive while slightly improving EPA.
These results indicate that persistence-aware reasoning improves the quality and stability of the scene representation without degrading standard perception metrics.

BeyondSight also improves downstream planning performance.
Planning error decreases from $0.61$ to $0.54$ L2$_{\text{avg}}$, while the collision rate decreases from $0.08\%$ to $0.07\%$.
Importantly, these gains are achieved without modifying the planner itself.
Instead, improvements arise from the richer scene representation produced by permanence-aware actor modeling.

Crucially, the permanence-aware model does not incur additional false positives under the official nuScenes evaluation protocol.
Standard mAP and NDS remain competitive with state-of-the-art baselines, demonstrating that maintaining persistent actor hypotheses does not negatively impact conventional benchmark performance.

\subsubsection{Object Permanence Performance.}
We next evaluate BeyondSight using the proposed observability-conditioned protocol on the nuScenes-Permanence benchmark.
Table~\ref{tab:observability_results} reports detection performance for observable actors, unobservable actors, and the full actor set.

\begin{table}[!htb]
\centering
\caption{Observability-conditioned detection and prediction performance on the nuScenes-Permanence validation set.}
\label{tab:observability_results}
\begin{tabular}{lcccc}
\toprule
Method & mAP$_\text{all}$ $\uparrow$ & NDS$_\text{all}$ $\uparrow$ & minADE$_\text{all}$ $\downarrow$ & EPA$_\text{all}$ $\uparrow$ \\
\midrule
 SparseDrive   & 0.389 & 0.514 & 0.642 & 0.457  \\
\rowcolor{highlight} BeyondSight & \textbf{0.413} & \textbf{0.519} & \textbf{0.582} & \textbf{0.481} \\ \midrule
Method & mAP$_\text{obs}$ $\uparrow$ & NDS$_\text{obs}$ $\uparrow$ & minADE$_\text{obs}$ $\downarrow$ & EPA$_\text{obs}$ $\uparrow$ \\
\midrule
 SparseDrive & 0.415 & 0.526 & \textbf{0.610} & 0.492 \\
\rowcolor{highlight} BeyondSight & \textbf{0.421} & \textbf{0.528} & 0.625 & \textbf{0.496} \\ \midrule
Method & mAP$_\text{unobs}$ $\uparrow$ & NDS$_\text{unobs}$ $\uparrow$ & minADE$_\text{unobs}$ $\downarrow$ & EPA$_\text{unobs}$ $\uparrow$ \\
\midrule
 SparseDrive & 0 & 0.255 & 0.615 & 0  \\
\rowcolor{highlight} BeyondSight & \textbf{0.249} & \textbf{0.306} & \textbf{0.479} & \textbf{0.285} \\
\bottomrule
\end{tabular}
\end{table}

BeyondSight substantially improves performance during occlusion.
The model achieves the highest mAP$_{\text{unobs}}$, improving over the strongest baseline by a large margin.
In contrast, performance on continuously observable actors remains nearly unchanged.
This demonstrates that permanence-aware supervision specifically improves reasoning about unobservable actors without degrading standard detection performance.
The aggregated metric mAP$_{\text{all}}$ further confirms improved holistic scene understanding.
By maintaining persistent hypotheses across observability gaps, BeyondSight produces a more complete representation of the dynamic environment.

\subsection{Ablation Study}

We conduct ablation experiments to analyze the contribution of the key components introduced in BeyondSight.
All ablations are evaluated on the nuScenes validation split using the same training protocol as the full model.
Table~\ref{tab:ablations} reports cumulative component additions under two annotation settings: standard nuScenes annotations and nuScenes-Permanence annotations. 

\begin{table}[!htb]
\centering
\caption{Ablation study of BeyondSight components under standard and permanence-aware annotations.}
\label{tab:ablations}
\begin{tabular}{lccccc}
\toprule
Configuration & mAP$_{\text{obs}}$ $\uparrow$ & mAP$_{\text{unobs}}$ $\uparrow$ & mAP$_{\text{all}}$ $\uparrow$ & L2$_{\text{avg}}$ $\downarrow$ & CR$_{\text{avg}}$ $\downarrow$ \\
\midrule
\multicolumn{6}{l}{\textit{Trained on standard nuScenes annotations}} \\
SparseDrive & 0.415 & 0.000 & 0.389 & 0.61 & 0.08 \\
+ Temporal prior & 0.417 & 0.000 & 0.390 & 0.60 & 0.08 \\
+ Posterior fusion & 0.413 & 0.000 & 0.388 & 0.61 & 0.08 \\
\midrule
\multicolumn{6}{l}{\textit{Trained on nuScenes-Permanence annotations}} \\
SparseDrive & 0.403 & 0.021 & 0.388 & 0.61 & 0.08 \\
+ Temporal prior & 0.406 & 0.126 & 0.399 & 0.58 & 0.08 \\
+ Posterior fusion & 0.415 & 0.194 & 0.405 & 0.56 & \textbf{0.07} \\
+ Unobs. supervision & 0.418 & 0.213 & 0.410 & 0.55 & \textbf{0.07} \\
\rowcolor{highlight} BeyondSight & \textbf{0.421} & \textbf{0.249} & \textbf{0.413} & \textbf{0.54} & \textbf{0.07} \\
\bottomrule
\end{tabular}
\end{table}

Under standard nuScenes annotations, unobservable actors are not explicitly supervised, and all configurations obtain zero mAP$_{\text{unobs}}$.
Adding the Temporal Prior Decoder slightly improves mAP$_{\text{obs}}$, mAP$_{\text{all}}$, and planning error, but does not enable detection of fully unobservable actors by itself.
With nuScenes-Permanence annotations, the same temporal prior provides a substantial gain on mAP$_{\text{unobs}}$, showing that temporal propagation is most effective when paired with permanence-aware supervision.

The Posterior Fusion Decoder further improves both observable and unobservable detection by reconciling propagated hypotheses with observation-conditioned queries.
Adding unobservable-actor supervision improves the persistence of occluded actors and reduces planning error.
The full BeyondSight model achieves the best overall performance, improving mAP$_{\text{unobs}}$ from 0.021 to 0.249 over the SparseDrive baseline trained on nuScenes-Permanence, while also improving mAP$_{\text{obs}}$, mAP$_{\text{all}}$, L2$_{\text{avg}}$, and CR$_{\text{avg}}$.

\subsection{Planning under Occlusion}

\begin{figure}[!htb]
\centering
\includegraphics[width=\linewidth]{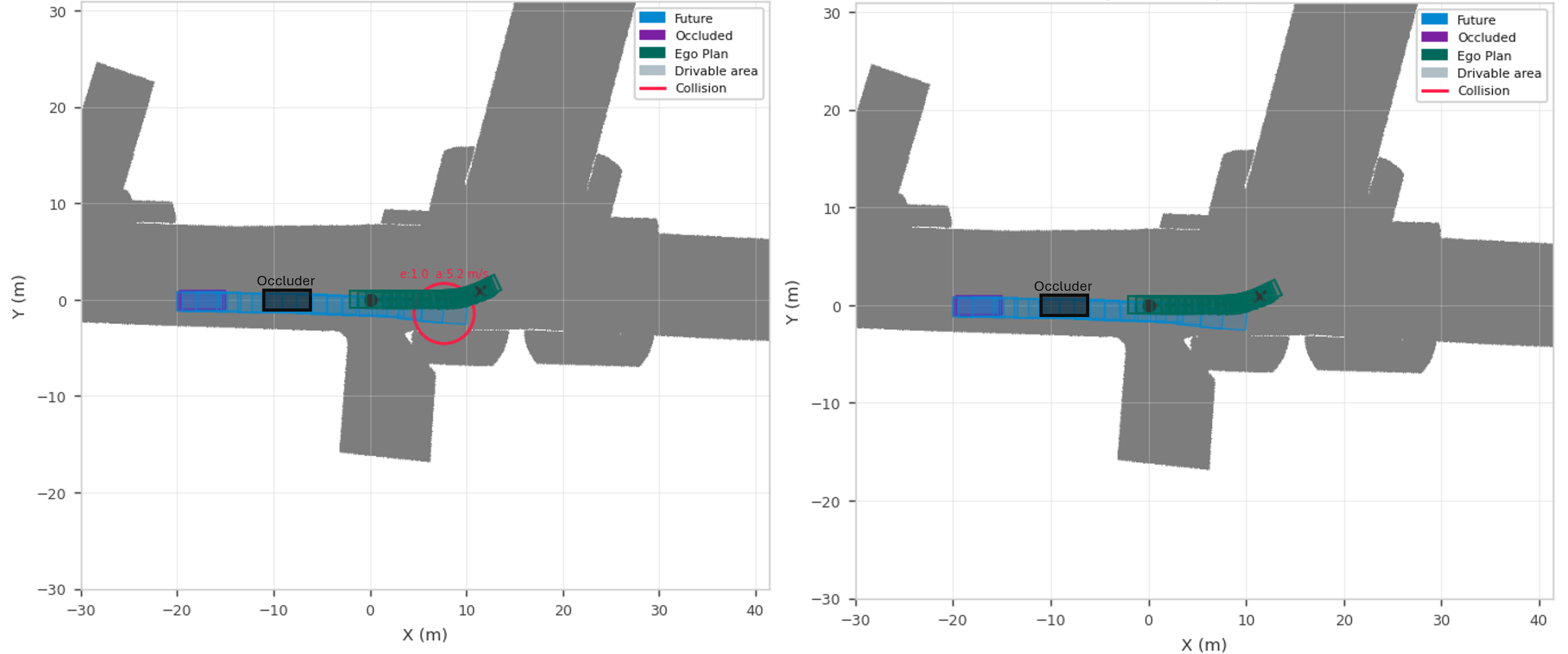}
\caption{Planning under full occlusion. Left: SparseDrive drops the occluded actor once visual evidence disappears, causing the ego plan to intersect the actor's ground-truth future trajectory. Right: BeyondSight maintains a persistent hypothesis for the hidden actor and produces a safer plan. The occluder, hidden actor, actor future, and ego plan are highlighted for clarity.}\
\label{fig:visualization}
\end{figure}

We qualitatively evaluate BeyondSight in scenarios involving partial observability and prolonged occlusions.
Figure~\ref{fig:visualization} shows a planning-relevant case where an actor becomes fully occluded by another vehicle.
SparseDrive drops the hidden actor once direct evidence disappears, causing the ego plan to intersect the actor's ground-truth future trajectory.
BeyondSight maintains a persistent hypothesis for the occluded actor and produces a plan with additional clearance.
This example illustrates how permanence-aware scene representations can provide useful context to downstream planning when hidden actors remain decision-relevant.

To quantify this behavior beyond a single qualitative example, we evaluate a high-occlusion subset containing 600 validation samples with occluded dynamic agents whose future trajectories approach the ego plan within 12\,m over a 4\,s horizon.
On this subset, SparseDrive degrades to 0.64 L2$_{\text{avg}}$ and 0.163 CR$_{\text{avg}}$, while BeyondSight achieves 0.56 L2$_{\text{avg}}$ and 0.08 CR$_{\text{avg}}$.
These results suggest that persistent actor hypotheses are most useful when unobservable actors remain relevant to the ego plan.

% ---------------------------------------------------------------
\section{Conclusion}
\label{sec:conclusion}

We presented BeyondSight, a permanence-aware end-to-end driving framework designed to operate reliably under partial observability.
Unlike existing systems that implicitly couple actor existence to instantaneous observations, BeyondSight maintains persistent actor hypotheses across occlusion intervals through temporally propagated and observation-conditioned updates.
We further introduced nuScenes-Permanence, an extension of the nuScenes benchmark that enables supervision and evaluation of actors during unobservable intervals through observability-conditioned metrics.
This benchmark provides the first systematic evaluation protocol for object permanence in end-to-end autonomous driving.

\textbf{Limitations.} BeyondSight's persistent hypotheses can become stale during long occlusions, leading to false persistence or localization drift when actors deviate from the learned temporal prior. nuScenes-Permanence unobservable annotations target occlusions with approximately continuous motion and may not capture abrupt hidden maneuvers such as sudden stops, turns, or intent changes. Future work should explore uncertainty-aware memory, simulation-based unobservable annotations with privileged labels, and multi-observer datasets for direct occlusion evaluation.

% ---- Bibliography ----
%
% BibTeX users should specify bibliography style 'splncs04'.
% References will then be sorted and formatted in the correct style.
%
\bibliographystyle{splncs04}
\bibliography{main}

% ---- Supplementary Material ----

\clearpage
\appendix

\begin{center}
    {\LARGE\bfseries Supplementary Material}
\end{center}

% --- Your supplementary content starts here ---

\section{Additional Implementation Details}

\subsubsection{Architecture.}
Our implementation builds on SparseDrive~\cite{sun2025sparsedrive} with a ResNet-50 backbone pretrained on ImageNet and an FPN producing four feature levels with strides $\{4,8,16,32\}$ and 256 channels. 
Multi-view inputs from six cameras are resized to $704 \times 256$. 
The transformer hidden dimension is $d=256$ with $8$ attention heads throughout.

\subsubsection{Perception Head.}
The 3D detection head follows the Sparse4D design with 900 anchor queries initialised via $k$-means clustering on the training set. 
The decoder contains six layers: the first processes single-frame features, while the remaining layers incorporate temporal context via a temporal attention over up to 600 propagated instances with confidence decay. 
Each layer applies graph attention, layer normalisation, deformable feature aggregation with twelve 3D keypoints per anchor (six fixed, six learnable), an FFN, and box/class refinement. 
An auxiliary visibility prediction head is trained with Focal Loss ($\gamma{=}0$, $\alpha{=}0.4$). 
Detection supervision uses Focal Loss for classification ($\gamma{=}2$, $\alpha{=}0.25$, weight $2.0$) and sparse 3D box regression (weight $0.25$). 

\subsubsection{Motion and Planning.}
The motion-planning module maintains an \texttt{InstanceQueue} storing the top-50 detected agents across $L{=}4$ frames with ego-motion compensation applied to anchors. 
Future agent trajectories are predicted over $T_{\text{fut}}{=}12$ steps ($\approx 6$\,s at 2\,Hz) with $K{=}6$ modes. 
Ego plans are predicted over $T_{\text{ego}}{=}6$ steps with six modes. 
The decoder applies three blocks of temporal attention, agent--agent attention, layer normalisation, and FFN before a final refinement stage. 
No cross-attention to map features is used. 
Motion and planning are supervised using Focal Loss for mode classification (weights $0.2$ and $0.5$) and L1 regression losses (weights $0.2$ and $1.0$). 
An additional L1 loss (weight $1.0$) supervises ego-status regression.

\subsubsection{Training.}
Models are trained in two stages on the nuScenes trainval split using 8 NVIDIA A100 GPUs. 
Stage 1 trains the sparse perception module for 100 epochs with a total batch size of 64, learning rate $4\times10^{-4}$, image-backbone learning-rate multiplier 0.5, and weight decay $10^{-3}$. 
Stage 2 jointly trains perception, motion prediction, and planning for 10 epochs with total batch size of 48, a learning rate of $3\times10^{-4}$, an image-backbone learning-rate multiplier of 0.1, and weight decay of $10^{-3}$.
Both stages use AdamW, cosine learning-rate scheduling with 500 warm-up iterations, and mixed precision (FP16). 
Data augmentation includes random resize ($[0.40,0.47]$), horizontal flip, and photometric distortion. 
Actors beyond 55\,m from the ego vehicle are filtered during training. 

\section{Additional Results}

\subsubsection{Additional Backbone Experiments.}
To evaluate the robustness of BeyondSight across backbone architectures, we conduct additional experiments using a ResNet-101 backbone. 
All training settings remain identical to the ResNet-50 configuration described in the main paper.
Table~\ref{tab:supp_backbone} compares SparseDrive and BeyondSight under both backbone configurations. 
This comparison evaluates whether the improvements from permanence-aware modeling persist when using a stronger visual encoder.
\begin{table}[!htb]
\centering
\small
\caption{Effect of backbone architecture on nuScenes validation performance. 
All other settings remain unchanged. $\uparrow$ indicates higher is better and $\downarrow$ indicates lower is better. *ResNet-101 baseline results reproduced using our reimplementation since official weights are not publicly available.}
\label{tab:supp_backbone}
\resizebox{1\linewidth}{!}{
\begin{tabular}{lcccccccc}
\toprule
Method 
& Backbone
& mAP$_\text{unobs}$ $\uparrow$
& mAP $\uparrow$
& NDS $\uparrow$
& AMOTA $\uparrow$
& minADE $\downarrow$
& L2$_{\text{avg}}$ $\downarrow$
& CR$_{\text{avg}}$ $\downarrow$ \\
\midrule
SparseDrive & ResNet-50 & 0 & 0.415 & 0.526 & 0.372 & 0.610 & 0.61 & 0.08 \\
\rowcolor{highlight} BeyondSight & ResNet-50 & \textbf{0.249} &\textbf{0.427} & \textbf{0.536} & \textbf{0.401} & \textbf{0.610} & \textbf{0.54} & \textbf{0.07} \\
\midrule
SparseDrive* & ResNet-101 & 0 & 0.494 & 0.585 & 0.501 & 0.605 & 0.60 & 0.08 \\
\rowcolor{highlight} BeyondSight & ResNet-101 & \textbf{0.274} & \textbf{0.506} & \textbf{0.599} & \textbf{0.539} & \textbf{0.595} & \textbf{0.53} & \textbf{0.07} \\
\bottomrule
\end{tabular}
}
\end{table}

\subsubsection{Extended Metrics Results.}
For brevity, several perception and prediction metrics were omitted from the main paper. 
This section reports the complete set of evaluation metrics for reference.
Table~\ref{tab:supp_full_mapping} reports the vectorized map prediction metrics.
Table~\ref{tab:supp_full_prediction} reports detailed trajectory prediction metrics for vehicles and pedestrians. 
We report Average Displacement Error (ADE), Final Displacement Error (FDE), Miss Rate (MR), and End-to-end Prediction Accuracy (EPA).

\begin{table}[!htb]
\centering
\small
\caption{Vectorized mapping results on the nuScenes validation set.}
\label{tab:supp_full_mapping}
\begin{tabular}{lcccc}
\toprule
Method
& AP$_{\text{ped}}$ $\uparrow$
& AP$_{\text{divider}}$ $\uparrow$
& AP$_{\text{boundary}}$ $\uparrow$
& mAP $\uparrow$ \\
\midrule
SparseDrive
& 49.9
& \textbf{57.0}
& 58.4
& 55.1 \\
\rowcolor{highlight} BeyondSight
& \textbf{52.1}
& 56.2
& \textbf{59.1}
& \textbf{55.8} \\
\bottomrule
\end{tabular}
\end{table}

\begin{table}[!htb]
\centering
\small
\caption{Full trajectory prediction metrics on the nuScenes validation set.}
\label{tab:supp_full_prediction}
\begin{tabular}{l|c|c|c|c}
\toprule
Method
& ADE (m) $\downarrow$
& FDE (m) $\downarrow$
& MR $\downarrow$
& EPA $\uparrow$ \\
 & (Car / Ped) & (Car / Ped) & (Car / Ped) & (Car / Ped) \\
\midrule
SparseDrive
& 0.61 / 0.72
& 0.99 / 1.07
& 0.14 / 0.14
& 0.49 / 0.41 \\
\rowcolor{highlight} BeyondSight
& \textbf{0.61} / \textbf{0.70}
& \textbf{0.96} / \textbf{1.01}
& \textbf{0.13} / \textbf{0.13}
& \textbf{0.50} / \textbf{0.43} \\
\bottomrule
\end{tabular}
\end{table}

\subsubsection{Latency Analysis.}
We measure inference latency on a single NVIDIA RTX 3090 GPU with batch size 1 and input resolution $704\times256$.
BeyondSight introduces a slight increase in computational overhead due to temporal propagation and permanence reasoning. 
Runtime decreases slightly from 6.1 FPS to 6.0 FPS (-1.8\%) compared to the SparseDrive baseline, indicating that permanence-aware modeling introduces minimal additional
computational cost.

\subsubsection{Occlusion-Duration Analysis.}

We additionally stratify object-permanence performance by occlusion duration.
For each unobservable actor, we compute the time since it was last observable and report detection quality, permanence recall, and false discovery rate.
Here, TPR$_{\text{unobs}}$ measures the fraction of unobservable actors that are maintained, while FDR$_{\text{unobs}}$ measures the fraction of maintained unobservable predictions that do not match a valid unobservable actor.
Table~\ref{tab:occlusion_duration} shows that performance degrades with longer occlusions: recall decreases and false persistence increases as hypotheses become stale.
This highlights long-duration occlusion as a key remaining challenge for permanence-aware models.

\begin{table}[!htb]
\centering
\caption{\textbf{Object permanence by occlusion duration.}
Longer occlusions reduce recall and increase false persistence.}
\begin{tabular}{lcccc}
\toprule
$T_{\text{occ}}$ 
& mAP$_{\text{unobs}}$ $\uparrow$ 
& NDS$_{\text{unobs}}$ $\uparrow$
& TPR$_{\text{unobs}}$ $\uparrow$ 
& FDR$_{\text{unobs}}$ $\downarrow$ \\
\midrule
0--2s & 0.307 & 0.401 & 0.62 & 0.85 \\
2--4s & 0.230 & 0.330 & 0.53 & 0.89 \\
4--6s & 0.219 & 0.226 & 0.35 & 0.98 \\
\bottomrule
\end{tabular}
\label{tab:occlusion_duration}
\end{table}

\section{nuScenes-Permanence Annotation Generation}
\label{sec:supp_permanence}

\begin{figure}[!htb]
  \centering
  \includegraphics[width=0.49\linewidth]{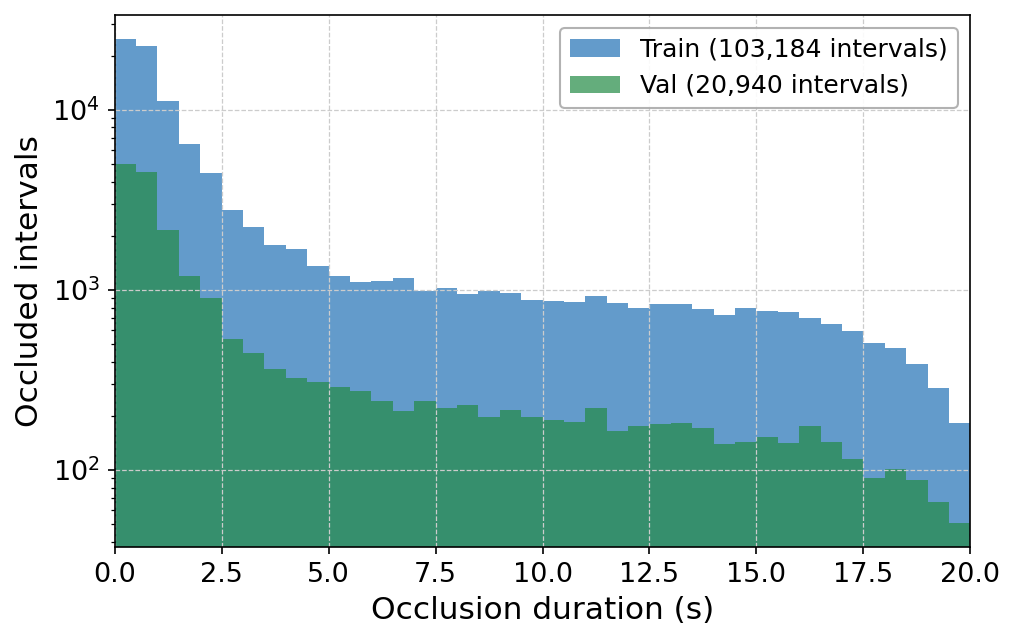}
  \hfill
  \includegraphics[width=0.49\linewidth]{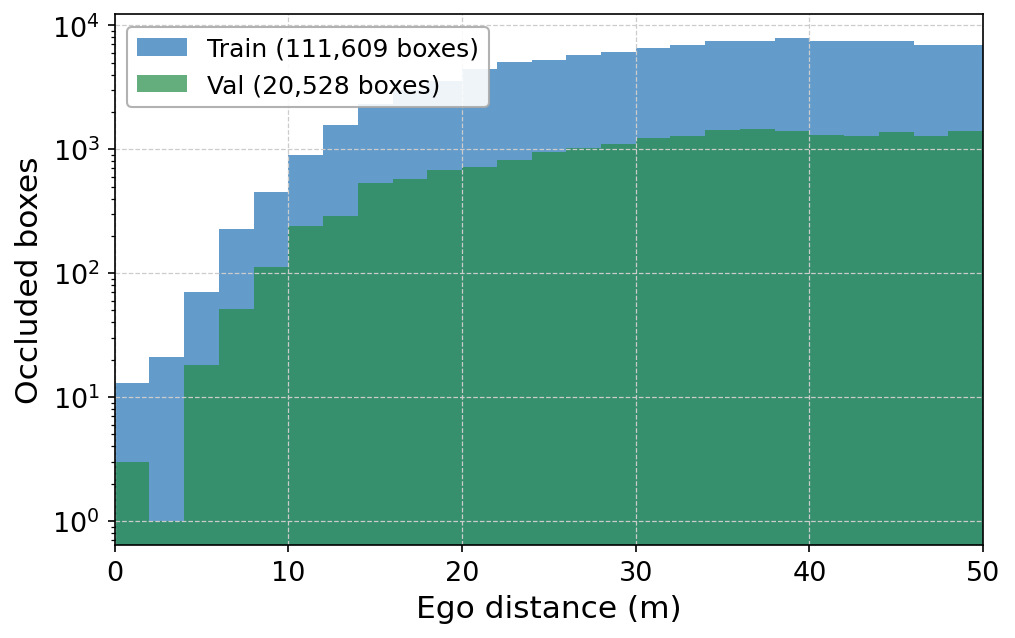}
  \caption{Distribution of full-occlusion durations (left) and ego distance (right) in nuScenes.}
  \label{fig:supp_occ_stats_main}
\end{figure}

This section provides implementation and validation details for the offline annotation generation procedure used to construct \textsc{nuScenes-Permanence}, together with additional analysis of the observability-conditioned evaluation protocol. 
We focus on details omitted from the main paper for brevity.
Figure~\ref{fig:supp_occ_stats_main} summarizes the prevalence of full observability gaps in nuScenes. 
The left panel reports the distribution of full-occlusion durations, while the right panel reports the corresponding ego-distance distribution. 
These statistics show that complete sensor absence occurs frequently across a broad range of distances and is not limited to rare edge cases.

% Figure~\ref{fig:supp_occlusion_stats} provides a finer breakdown by dynamic class. We report both the frequency of full sensor-absence events and the cumulative time spent under complete unobservability. Prolonged observability gaps occur across multiple actor categories, confirming that persistence under occlusion is a structural property of the dataset rather than an isolated failure mode.

% \begin{figure}[!htb]
%   \centering
%   \includegraphics[width=0.8\linewidth]{example-image}
%   \caption{Per-class breakdown of observability gaps. \textbf{Placeholder:} include histogram of gap durations by class and cumulative unobservable time by class.}
%   \label{fig:supp_occlusion_stats}
% \end{figure}

\subsubsection{Annotation Generation Process.}
To supervise persistence through observability gaps, annotations must remain defined when actors are physically present but fully unobservable. Since such intervals are not consistently retained under the standard nuScenes filtering rules, we construct additional annotations offline using a trajectory completion procedure.

For each actor track, we identify contiguous intervals of missing observability and estimate the latent actor state during those intervals. Each generated state is assigned a provenance label
\[
\tau_t^i \in \{\texttt{observable},\texttt{zero-point},\texttt{interpolated},\texttt{extrapolated}\},
\]
which is stored and later used for analysis and evaluation.

\emph{(1) Zero-point boxes.}
Some annotated boxes remain temporally valid despite having zero LiDAR or radar support at a given timestamp. Rather than discarding these boxes, we retain them as direct supervision for actors that are heavily occluded but still present in the scene.

\emph{(2) Temporary gaps: interpolation.}
For actors that disappear and later reappear within the scene horizon, we reconstruct the intermediate states by fitting a motion prior between the last observable state and the first subsequent observable state. We use a constant turn-rate and acceleration model in SE(2), constrained by the endpoint positions, headings, and velocities:
\[
\mathbf{x}_t = (x_t, y_t, \theta_t, v_t).
\]
Let $t_a$ and $t_b$ denote the last observable and first re-observed timestamps surrounding a gap. Intermediate states for $t \in (t_a, t_b)$ are generated by solving for a smooth kinematic trajectory consistent with the endpoint conditions. 
% \textbf{Placeholder:} add exact interpolation equations or optimization objective if you want full reproducibility.

\emph{(3) Terminal gaps: extrapolation.}
For actors that do not reappear before the scene ends, future states are estimated using a pretrained trajectory forecasting prior. This prior predicts short-horizon future motion from the last observable history. For stationary or near-stationary actors, or when the forecast confidence is below a threshold, we instead apply analytic kinematic propagation to avoid implausible drift:
\[
\mathbf{x}_{t+1} = f_{\text{kin}}(\mathbf{x}_t).
\]
Table~\ref{tab:permanence_stats} summarizes the composition of the resulting annotation set and the relative contribution of each trajectory completion mechanism.
\begin{table}[!ht]
\centering
\caption{Statistics of the nuScenes-Permanence annotation extension. 
Original annotations correspond to standard nuScenes labels. 
Additional annotations arise from retaining zero-point boxes and completing trajectories during unobservable intervals.}
\label{tab:permanence_stats}
\begin{tabular}{lcc}
\toprule
Annotation Type & Boxes & Fraction \\
\midrule
Original nuScenes annotations & 932k & 70\% \\
Zero-point (sensor-free) boxes & 266k & 20\% \\
Interpolated trajectory states & 11k & 1\% \\
Extrapolated terminal states & 119k & 9\% \\
\midrule
Total (nuScenes-Permanence) & 1.33M & 100\% \\
\bottomrule
\end{tabular}
\end{table}

\subsubsection{Forecasting Model.}

The motion forecasting model is trained on the nuScenes \texttt{v1.0-trainval} observable trajectory annotations.
The model is adapted from SparseDrive~\cite{sun2025sparsedrive} but removes the planning head and replaces the detection and mapping heads with ground truth object and map information.
The model is trained on all detection class categories.
The model is trained to predict 6\,s futures, sampled at 2\,Hz.
Map context is encoded from six polyline types---lanes, stop signs, road edges, road lines, crosswalks, and speed bumps---within a \(150\,\text{m}\) radius, centered \(30\,\text{m}\) ahead of the focal agent.
The \(z\)-axis and physical size attributes are masked from the input, reducing the task to bird's-eye-view trajectory prediction.
The model uses a ResNet-50 image encoder and 4-layer transformer decoder (\(d{=}128\), 8 heads) and trained with the same hyperparameter settings as BeyondSight.

Generated trajectories are filtered for physical consistency. 
In particular, we reject or clip trajectories that exhibit implausible velocity spikes, abrupt heading changes, or large off-drivable-area deviations for vehicle classes.
Bounding-box dimensions and semantic class labels are retained from the nearest reliable observable state unless evidence suggests a valid state transition.
The annotation model is used strictly offline during dataset construction and is never used at inference time.

\subsubsection{Validation of New Annotations.}
To assess the quality of the generated annotations, we perform a holdout reconstruction study on observable trajectory intervals, Table~\ref{tab:pseudo_label_quality}. 
For each selected interval, we remove the actor states from the annotation sequence, reconstruct them using the same interpolation or extrapolation procedure used for nuScenes-Permanence, and compare the reconstructed states to the original nuScenes ground truth. 
Errors are stratified by annotation provenance, object class, occlusion duration, and motion pattern. 
We then prevalence-weight the strata according to the distribution of generated unobservable labels used in evaluation. 
This provides an empirical estimate of the uncertainty introduced by the annotation-generation process.

\begin{table}[!htb]
\centering
\caption{\textbf{Label quality analysis.}
Top: all labels, including zero-point GT.
Bottom: generated labels only, excluding zero-point.}
\label{tab:pseudo_label_quality}
\footnotesize
\begin{tabular}{lcccc}
\toprule
\textbf{Label Subset} & \textbf{Labels} & \textbf{L2 err. (m) $\downarrow$} & \textbf{BEV IoU $\uparrow$} & \textbf{Hit Rate $\uparrow$} \\
& \% & ($\mu$, $P_{90}$) & ($\mu$, $P_{10}$) & ($<$2m, $<$1m) \\
\midrule \midrule
\textbf{All} & 1.00 & -- & -- & -- \\
\midrule
Zero-point & 0.67 & -- & -- & -- \\
Extrapolated & 0.30 & 0.56 / 1.15 & 0.72 / 0.22 & 0.94 / 0.90 \\
Interpolated & 0.03 & 0.24 / 0.60 & 0.88 / 0.47 & 0.99 / 0.96 \\
\midrule \midrule
% \textbf{Generated} & 100 & 0.45 / 2.0 & \placeholder{} / \placeholder{} & \placeholder{} / \placeholder{} \\
\textbf{Generated} & 1.00 & 0.50 / 1.04 & 0.77 / 0.36 & 0.95 / 0.92 \\
\midrule
Vehicle & 0.57 & 0.73 / 1.92 & 0.80 / 0.42 & 0.92 / 0.88 \\
Pedestrian & 0.21 & 0.37 / 0.95 & 0.62 / 0.30 & 0.97 / 0.91 \\
Movable & 0.23 & 0.07 / 0.18 & 0.81 / 0.41 & 1.00 / 1.00 \\
\midrule
0--2s & 0.43 & 0.14 / 0.33 & 0.84 / 0.52 & 0.99 / 0.97 \\
2--4s & 0.33 & 0.52 / 1.41 & 0.74 / 0.32 & 0.94 / 0.89 \\
4--6s & 0.24 & 1.11 / 3.22 & 0.68 / 0.30 & 0.89 / 0.84 \\
\midrule
Stationary & 0.53 & 0.05 / 0.14 & 0.90 / 0.67 & 1.00 / 1.00 \\
Straight & 0.31 & 1.09 / 2.92 & 0.61 / 0.30 & 0.88 / 0.79 \\
Turning & 0.16 & 0.85 / 1.95 & 0.62 / 0.30 & 0.92 / 0.87 \\
\bottomrule
\end{tabular}
\end{table}

The results show that reconstruction quality is highest for retained zero-point boxes and interpolation, and degrades with longer extrapolation horizons and dynamic motion. 
This behavior motivates the adaptive matching tolerance used in observability-conditioned evaluation.
Reconstruction error increases smoothly with horizon and remains bounded over the occlusion durations most commonly observed in nuScenes. 
These results indicate that the generated supervision reflects controlled motion uncertainty, making it suitable for training and evaluation under temporary unobservability.

\begin{figure}[!htb]
  \centering
  \includegraphics[width=0.8\linewidth]{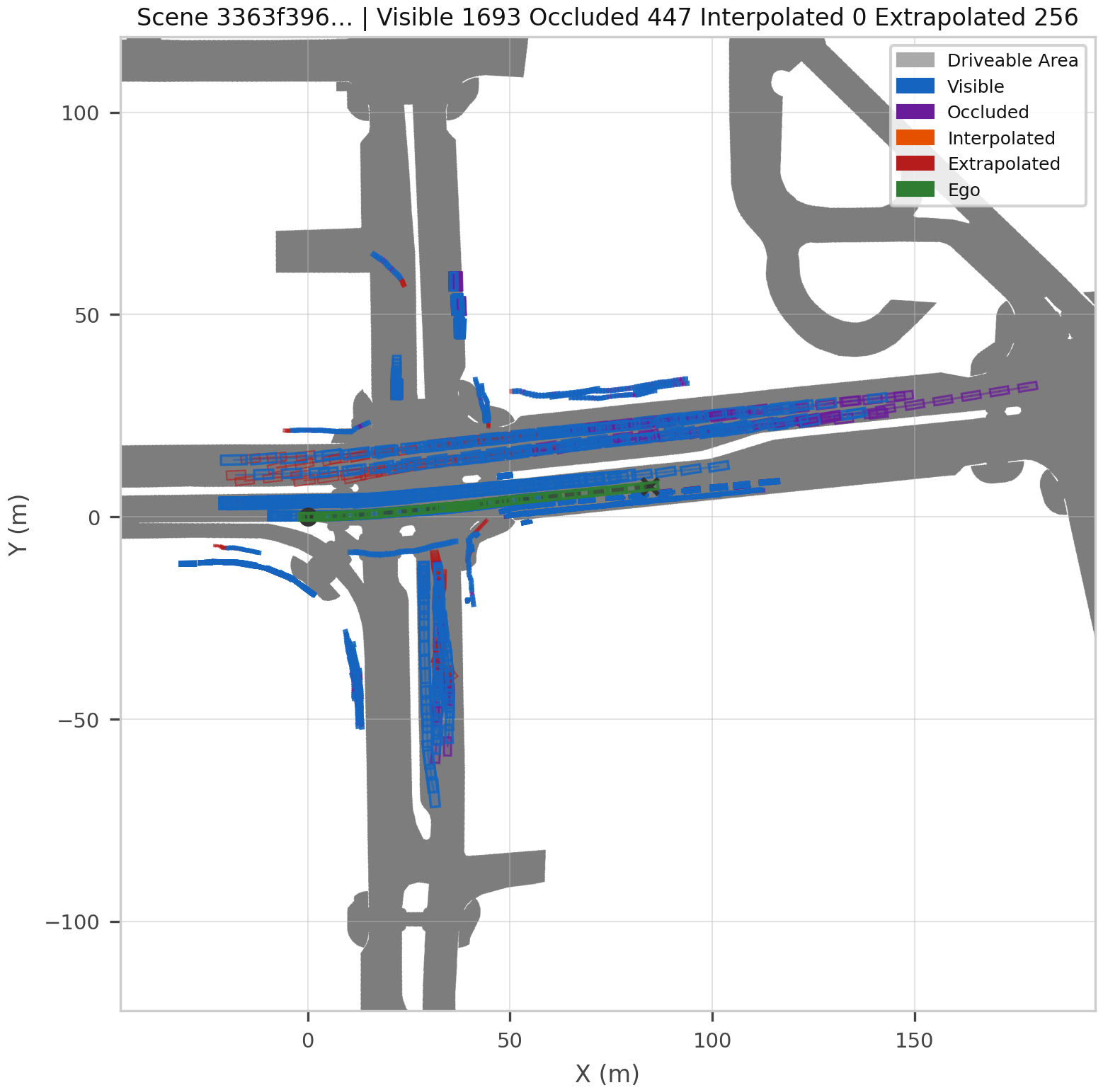}
  \caption{Examples of generated annotations including full unobservability for a single scene. Boxes are colored according to their observability status and dataset generation method.}
  \label{fig:supp_annotation_vis}
\end{figure}

Figure~\ref{fig:supp_annotation_vis} shows representative examples of retained zero-point boxes, interpolated segments, and extrapolated terminal segments across a range of traffic scenarios, including urban intersections, turns, and highway scenes. In each case, the completed annotations remain temporally smooth and geometrically consistent with the road layout.

\section{Evaluation Metrics and nuScenes-Permanence}
\label{sec:supp_eval}

For clarity, Table~\ref{tab:metric_definitions} summarizes the evaluation metrics used throughout the paper.
Standard nuScenes detection metrics include mAP, mATE, mASE, mAOE, mAVE, mAAE, and NDS.
Tracking is evaluated using AMOTA, AMOTP, recall, and identity switches.
Motion prediction is evaluated using minADE, minFDE, Miss Rate, and EPA, while planning is evaluated using open-loop trajectory error and collision rate.
For nuScenes-Permanence, these metrics are further partitioned into observable, unobservable, and all-actor subsets using the observability-conditioned protocol described below.

\begin{table}[!htb]
\centering
\caption{Summary of evaluation metrics.}
\label{tab:metric_definitions}
\resizebox{\linewidth}{!}{
\begin{tabular}{lll}
\toprule
Category & Metric & Definition \\
\midrule
Detection 
& mAP $\uparrow$ 
& Mean average precision over 3D detection classes. \\
& mATE $\downarrow$ 
& Mean average translation error of matched detections. \\
& mASE $\downarrow$ 
& Mean average scale error of matched detections. \\
& mAOE $\downarrow$ 
& Mean average orientation error of matched detections. \\
& mAVE $\downarrow$ 
& Mean average velocity error of matched detections. \\
& mAAE $\downarrow$ 
& Mean average attribute error of matched detections. \\
& NDS $\uparrow$ 
& nuScenes detection score combining mAP and detection error terms. \\
\midrule
Tracking 
& AMOTA $\uparrow$ 
& Average multi-object tracking accuracy over recall thresholds. \\
& AMOTP $\downarrow$ 
& Average multi-object tracking precision over recall thresholds. \\
& Recall $\uparrow$ 
& Fraction of ground-truth tracks or detections recovered. \\
& IDS $\downarrow$ 
& Number of identity switches in tracking. \\
\midrule
Motion prediction 
& minADE $\downarrow$ 
& Minimum average displacement error over predicted trajectory modes. \\
& minFDE $\downarrow$ 
& Minimum final displacement error over predicted trajectory modes. \\
& MR $\downarrow$ 
& Miss rate under the forecasting displacement threshold. \\
& EPA $\uparrow$ 
& End-to-end prediction accuracy for detection-conditioned forecasting. \\
\midrule
Planning 
& L2$_{1s,2s,3s}$ $\downarrow$ 
& Ego trajectory L2 error at 1, 2, and 3 second horizons. \\
& L2$_{\text{avg}}$ $\downarrow$ 
& Average ego trajectory L2 error across planning horizons. \\
& Col$_{1s,2s,3s}$ $\downarrow$ 
& Ego collision rate at 1, 2, and 3 second horizons. \\
& CR$_{\text{avg}}$ $\downarrow$ 
& Average ego collision rate across planning horizons. \\
\midrule
Permanence 
& mAP$_{\text{obs}}$ $\uparrow$ 
& Detection mAP evaluated only on observable actors. \\
& mAP$_{\text{unobs}}$ $\uparrow$ 
& Detection mAP evaluated only on unobservable actors. \\
& mAP$_{\text{all}}$ $\uparrow$ 
& Detection mAP evaluated over observable and unobservable actors. \\
& NDS$_{\text{obs/unobs/all}}$ $\uparrow$ 
& NDS computed on observable, unobservable, or all actors. \\
& minADE$_{\text{obs/unobs/all}}$ $\downarrow$ 
& Forecasting error computed on observable, unobservable, or all actors. \\
& EPA$_{\text{obs/unobs/all}}$ $\uparrow$ 
& End-to-end prediction accuracy for each observability subset. \\
\bottomrule
\end{tabular}}
\end{table}

The main paper introduces the observability-conditioned evaluation protocol. 
Here we summarize the implementation details used in our experiments.

\subsubsection{Observability-Conditioned Evaluation.}
At each timestamp the ground-truth actor set is partitioned into observable and unobservable subsets
\[
G_{\mathrm{obs}}, \qquad G_{\mathrm{unobs}}.
\]
Evaluation is performed by selecting a target set $G$ and defining the complementary subset as an ignore set $I$. 
Predictions are first matched to $G$ using the standard nuScenes class-aware distance matching rule. 
Remaining predictions that match any element of $I$ are removed before false-positive accumulation. 
This symmetric ignore rule prevents predictions corresponding to the complementary subset from being incorrectly penalized.

We report the following evaluation subsets:
\begin{align}
\mathrm{mAP}_{\mathrm{obs}} &: \text{performance on } G_{\mathrm{obs}}, \\
\mathrm{mAP}_{\mathrm{unobs}} &: \text{performance on } G_{\mathrm{unobs}}, \\
\mathrm{mAP}_{\mathrm{all}} &: \text{performance on } G_{\mathrm{obs}} \cup G_{\mathrm{unobs}} .
\end{align}

\begin{table}[!htb]
  \centering
  \caption{Observability-conditioned detection metrics.}
  \label{tab:det_metrics}
  \begin{tabular}{lcccc}
  \toprule
  \textbf{Metric} & \textbf{GT} & \textbf{Ignore} & \textbf{TP} & \textbf{FP} \\
  \midrule
  mAP 
      & $\mathcal{G}_{\text{obs}}$
      & $\varnothing$
      & $M_{\text{obs}}$
      & $M_{\emptyset} \cup M_{\text{unobs}}$ \\
  \midrule
  $\mathrm{mAP}_{\text{obs}}$
      & $\mathcal{G}_{\text{obs}}$
      & $\mathcal{G}_{\text{unobs}}$
      & $M_{\text{obs}}$
      & $M_{\emptyset}$ \\
  $\mathrm{mAP}_{\text{unobs}}$
      & $\mathcal{G}_{\text{unobs}}$
      & $\mathcal{G}_{\text{obs}}$
      & $M_{\text{unobs}}$
      & $M_{\emptyset}$ \\
  $\mathrm{mAP}_{\text{all}}$
      & $\mathcal{G}_{\text{obs}} \cup \mathcal{G}_{\text{unobs}}$
      & $\varnothing$
      & $M_{\text{obs}} \cup M_{\text{unobs}}$
      & $M_{\emptyset}$ \\
  \bottomrule
  \end{tabular}
\end{table}

\subsubsection{Adaptive Matching Threshold.}
Evaluation against extrapolated ground truth introduces additional localization uncertainty. 
To avoid conflating model error with extrapolation drift, we augment the standard nuScenes matching threshold with an adaptive tolerance:

\begin{equation}
d_{\mathrm{match}}(v,a,t,\ell)
=
d_{\mathrm{nuScenes}}(\ell)
+
\hat{d}_\ell(v,a,t),
\end{equation}

where $t$ denotes the duration of unobservability and $\ell$ the object class.
The additional tolerance is modeled as

\begin{equation}
\hat{d}_\ell(v,a,t)
=
\alpha_\ell t
+
\beta_\ell v t
+
\gamma_\ell a t^2,
\end{equation}

capturing drift due to velocity and acceleration uncertainty during occlusion, where $v$ and $a$ denote the actor's ground truth speed and acceleration magnitude at the last observable state.
The coefficients $(\alpha_\ell,\beta_\ell,\gamma_\ell)$ are estimated using non-negative 90th-percentile quantile regression on approximately $1.5\times10^6$ extrapolated box–horizon pairs generated by masking observable nuScenes annotations.

\begin{table}[!htb]
\centering
\caption{Adaptive matching coefficients.}
\label{tab:adaptive_coeffs}
\begin{tabular}{lccc}
\toprule
Class & $\alpha$ & $\beta$ & $\gamma$ \\
\midrule
Vehicle    & 0.0568 & 0.1962 & 0.2133 \\
Cyclist    & 0.1023 & 0.1861 & 0.2266 \\
Pedestrian & 0.2641 & 0.1457 & 0.1774 \\
\bottomrule
\end{tabular}
\end{table}

As $t \rightarrow 0$, $\hat{d}_\ell \rightarrow 0$, recovering the official nuScenes evaluation protocol for observable actors.

% \begin{figure}[!htb]
%   \centering
%   \includegraphics[width=0.8\linewidth]{example-image}
%   \caption{Extrapolation error growth with quantile fit.}
%   \label{fig:supp_error_growth}
% \end{figure}

\section{Limitations and Failure Cases}

While BeyondSight improves reasoning under partial observability, several limitations remain. 
First, very long occlusion intervals ($>6$\,s) can lead to drift in propagated hypotheses when actor motion deviates from the assumed dynamics. 
Second, abrupt behavioral changes (e.g., sudden stops or turns after occlusion) remain difficult to anticipate because the model relies primarily on historical motion priors. 
Third, persistence may occasionally maintain low-confidence hypotheses for actors that have permanently exited the scene, which can introduce minor localization noise until the hypothesis is pruned by downstream confidence filtering. 
Addressing these challenges may require stronger interaction-aware forecasting or explicit uncertainty modeling during long observability gaps.

\end{document}